\documentclass[11pt,a4paper,dvipsnames]{article}
\usepackage{url}
\usepackage{comment}
\usepackage[above,below]{placeins}

\usepackage[]{EMNLP2023}
\usepackage{times}
\usepackage{latexsym}

\usepackage[T1]{fontenc}

\usepackage[utf8]{inputenc}

\usepackage{microtype}

\usepackage{inconsolata}

\usepackage{xcolor}
\usepackage{graphicx}
\usepackage{hyperref}
\usepackage{booktabs}
\usepackage{amsmath}
\usepackage{cleveref}
\usepackage{csquotes}
\usepackage{siunitx}
\usepackage{subcaption}
\usepackage{inconsolata}
\usepackage{soul}

\DeclareSIUnit{\million}{M}

\title{Scaling Laws of Decoder-Only Models on the Multilingual Machine Translation Task}

\author{Gaëtan Caillaut \and Raheel Qader \and Mariam Nakhlé \\ \and \textbf{Jingshu Liu} \and \textbf{Jean-Gabriel Barthélemy} \\
    Lingua Custodia, Paris, France \\
    \texttt{firstname.name@linguacustodia.com}
}
\date{April 2024}

\begin{document}

\maketitle

\begin{abstract}
    Recent studies have showcased remarkable capabilities of decoder-only models in many NLP tasks, including translation. Yet, the machine translation field has been largely dominated by encoder-decoder models based on the Transformer architecture. As a consequence, scaling laws of encoder-decoder models for neural machine translation have already been well studied, but decoder-only models have received less attention.
    This work explores the scaling laws of decoder-only models on the multilingual and multidomain translation task. We trained a collection of six decoder-only models, ranging from 70M to 7B parameters, on a sentence-level, multilingual and multidomain dataset. We conducted a series of experiments showing that the loss of decoder-only models can be estimated using a scaling law similar to the one discovered for large language models, but we also show that this scaling law has difficulties to generalize to \textit{too large models} or to a different data distribution. We also study different scaling methods and show that scaling the depth and the width of a model lead to similar test loss improvements, but with different impact on the model's efficiency.
\end{abstract}

\section{Introduction}

Most modern machine translation systems are based on Transformers~\citep{vaswani2017attention}, with an encoder-decoder architecture. Despite the tremendous advances made possible with the release of open-source decoder-only Large Language Models (LLMs)~\citep{jiang2023mistral,biderman2023pythia,touvron2023llama}, most NLP tasks still rely on encoder-decoder models. Based on the statistics obtained from the WMT23 shared task on general machine translation~\citep{kocmi2023findings}, 16 out of the 17 participants submitted a system based on an encoder-decoder model. Yet, recent studies show that decoder-only models can achieve comparable results~\citep{gao2022encoder,fu2023decoder}, or even surpass state-of-the-art encoder-decoder systems, when properly fine-tuned~\citep{xu2023paradigm}.
Moreover, the decoder-only architecture is easier to train on massive amounts of data as one can simply concatenate documents and feed as much relevant data as possible into the model during training ; while encoder-decoder models requires either to pad the inputs or need rely on complex masking strategies~\citep{raffel2020exploring} to combine multiple inputs in the same sample.

Furthermore, the decoder architecture is much more flexible than the encoder-decoder architecture as decoders treat all tokens similarly, while encoder-decoders make a distinction between input (source) tokens and output (target) tokens, which are processed, respectively, by the encoder and the decoder.
As a consequence, it is more tedious to apply complex \textit{self-reasoning} mechanisms, such as chain-of-thought~\citep{wei2022chain}, or to interface it with external tools~\citep{schick2024toolformer}, because the outputs of such method (the \textit{reasoning} process) should, preferably, be treated as inputs of the model. For the same reasons, it is much more computationally expensive to rely on an encoder-decoder for conversational purposes, making this architecture less efficient for modern workflows such as iterative translation. Indeed, at each round (the user's query and the system's answer) should be appended to the input side, and reprocessed by the encoder for the next round. Decoder-only models support it by design, without needing to recompute the representation of the ever-growing inputs.
While we do not explore these directions in this work, we do leverage the flexibility of the decoder architecture to include input-or-output parameters. As we are tackling the multilingual and multidomain machine translation task, the model needs input tokens to represent the language direction and the domain. We propose to train the model to predict the source language and the domain so that, during inference, they can be seamlessly predicted or provided by the user.


Generally speaking, decoder-only models simply expect the input to be the whole discussion and process it in a single forward step. Causal masking enable efficient caching of already computed keys and values so inference is much cheaper. The main downside is that the quality of the input representation might be inferior, as input tokens can attend only on past tokens. But it should not be a major issue, as generated tokens attend to the whole past sequence, they do have access to the same quantity of information as with an encoder-decoder model. In addition, previous work propose to update the attention mask so that input tokens can attend to all input tokens while generated tokens can attend only on past tokens~\citep{tay2022ul2,raffel2020exploring}.

For all these reasons, we would like to embrace the decoder architecture for machine translation, even if it seems to be the exclusive preserve of encoder-decoder models. The flexibility and the simpler training setup of decoders should make them both more suitable and efficient for most real world applications, and the decoder architecture is more appropriate to answer the ever-growing demand for iterative, interactive and machine assisted translation workflow.
To this aim, we study the scaling laws of neural machine translation models under different settings. Our contributions are as follow:
\begin{itemize}
    \item \textbf{We show that decoder-only models for translation follow the same scaling law as LLM}
    \item \textbf{Scaling laws do not scale uniformly across directions and domains and do not generalize well to other directions or domains}
    \item \textbf{Scaling width-wise and depth-wise yield similar improvements, but the former is more efficient}
    \item \textbf{We discovered a critical issue related to the packing of training samples in batches and propose a solution to fix it}
\end{itemize}


\section{Background}

As the model size, data requirement, and training costs of language models rise, it quickly becomes extremely important to estimate the \textit{right} training configuration for a given training budget --- expressed in number of floating point operations (FLOP) --- required to train the model. \citet{kaplan2020scaling} discovered a power law relationship between the loss of a language model and its number of parameters, and that larger models perform better given the same amount of data. Even though certain work in this area shows that larger models, indeed, tend to be more powerful, more recent studies show that other parameters must be taken into account as well. For instance, the \textit{Chinchilla} scaling law~\citep{hoffmann2022training} shows that model and dataset sizes are loosely tied and need to be scaled equally. In other words, even if increasing only the model size will most likely improve its performances, the compute-optimal solution often requires to also increase the quantity of training data, while preserving the same training cost. These findings had a great impact on LLM research, as researchers stopped increasing blindly the size of their models, in favor of more data, when it was necessary. For instance, the 176B BLOOM~\citep{le2022bloom} model would probably have been trained very differently if this study was released sooner (or not at all). As stated in the paper, \enquote{in light of revised scaling laws published during training, we decided to train the large models for an additional 25 billion tokens on repeated data}, the authors discovered that training such a big model was sub-optimal given the quantity of data they had. As a consequence, many researchers started to work on the collection of large, high quality datasets~\citep{nguyen2023culturax,penedo2023refinedweb} or on means to enhance existing datasets~\citep{sorscher2022beyond,tirumala2024d4}.

Most of these scaling laws studies focus exclusively on causal generative language models. While it's likely that many of these findings could apply to translation models, the differences between the two tasks cannot be taken for granted. Translation is way more strict than causal language modeling since the model has to take into account each information in the source and precisely generate the target sentence without adding or omitting any information. Hence, many studies have naturally emerged to observe the scaling behavior of translation models~\citep{gordon2021data,fernandes2023scaling,ghorbani2021scaling}. Yet, most (if not all) of these works focused on encoder-decoder models. For instance, \citet{gordon2021data,fernandes2023scaling} showed that, when the encoder and the decoder are scaled proportionally, the model's performances follow a power-law similar to the observation made on language models. \citet{ghorbani2021scaling} tackle the problem in a different setup, and propose to scale the encoder and the decoder individually. They show that encoder-scaling and decoder-scaling affect the model's performances differently, and they propose a new formula describing the scaling behavior of the cross-entropy loss as a bivariate function of encoder and decoder size. They found out that scaling decoder is, according to their experiments, always more beneficial, in terms of cross-entropy loss performance, than scaling the encoder.

Recently, \citet{tower_llm_2024} introduced the TowerInstruct, an LLM based on a decoder architecture (LLama 2~\citep{touvron2023llama}) finetuned to handle several translation tasks. They show that a properly fine-tuned LLM can perform translation better than state-of-the-art models. But the most promising aspect of this work is the inherent capacity of LLM to handle different tasks. They finetuned TowerInstruct so it can, for instance, clean source sentences before translating them, follow terminological constraints or respect a given level of language. However, this work is still empirical and we do not know, yet, the limits of such models.
Inspired by the performances of TowerInstruct, a decoder-based machine translation system, we study, in the following, the scaling behavior of decoder-based machine translation models trained from scratch. To this aim, we fit multiple scaling laws to see if translation models follow the same scaling laws as language modeling models (such as the \textit{Chinchilla} law) or if they follow their own task-specific law.

\section{Training methodology}

We present in this section all details related to the training of our six models.

\subsection{Data}\label{subsec-data}

To conduct our experiments, we collected many bilingual data from public repositories (CCMatrix~\citep{schwenk2019ccmatrix}, WikiMatrix~\citep{schwenk2019wikimatrix}, UN Parallel Corpus~\citep{ziemski2016united}, Paracrawl~\citep{banon2020paracrawl} and Europarl~\citep{koehn2005europarl}). We also included a subset of an in-house proprietary dataset collected over time, as well as a small portion of financial documents in order to observe the scaling behavior on domain-specific data.
An overview of the dataset distribution is given in \Cref{tab-data-overview}. The financial data is divided into 8 sub-domains, which are described in \Cref{sec-full-data-distribution}.
The data is made of bilingual texts with one sample being one sentence pair.

\begin{table}[ht]
    \centering
    \begin{tabular}{llS[round-mode=places, round-precision=2, table-format=5.2, fixed-exponent=6, table-omit-exponent, table-alignment-mode=format, table-number-alignment=right, table-alignment=right]S[round-mode=places, round-precision=2, table-format=5.2, fixed-exponent=6, table-omit-exponent, table-alignment-mode=format, table-number-alignment=right, table-alignment=right]}
        \toprule
        Pair            & Domain  & {Sentences}              & {Tokens}                   \\
        \midrule
        ende            & general & \SI{46528535}{\million}  & \SI{2694164197}{\million}  \\
                        & finance & \SI{1290011}{\million}   & \SI{65926309}{\million}    \\
        enes            & general & \SI{51875735}{\million}  & \SI{3525208003}{\million}  \\
                        & finance & \SI{1336689}{\million}   & \SI{71481185}{\million}    \\
        enfr            & general & \SI{81389930}{\million}  & \SI{5430769538}{\million}  \\
                        & finance & \SI{8289408}{\million}   & \SI{494467216}{\million}   \\
        enit            & general & \SI{26214900}{\million}  & \SI{1657578525}{\million}  \\
                        & finance & \SI{732558}{\million}    & \SI{36167125}{\million}    \\
        ennl            & general & \SI{42741603}{\million}  & \SI{2057809691}{\million}  \\
                        & finance & \SI{1360229}{\million}   & \SI{63963157}{\million}    \\
        enpt            & general & \SI{42024152}{\million}  & \SI{2086617605}{\million}  \\
                        & finance & \SI{608780}{\million}    & \SI{22554913}{\million}    \\
        ensv            & general & \SI{46353741}{\million}  & \SI{2180637803}{\million}  \\
                        & finance & \SI{241534}{\million}    & \SI{9677142}{\million}     \\
        frde            & general & \SI{23596257}{\million}  & \SI{1470677693}{\million}  \\
                        & finance & \SI{1461153}{\million}   & \SI{72918353}{\million}    \\
        fres            & general & \SI{32902641}{\million}  & \SI{2731787517}{\million}  \\
                        & finance & \SI{481619}{\million}    & \SI{23391065}{\million}    \\
        frit            & general & \SI{28016216}{\million}  & \SI{1845841293}{\million}  \\
                        & finance & \SI{1104489}{\million}   & \SI{61633958}{\million}    \\
        frnl            & general & \SI{31937498}{\million}  & \SI{2034743248}{\million}  \\
                        & finance & \SI{619198}{\million}    & \SI{29180257}{\million}    \\
        \midrule
        \textbf{Total:} & general & \SI{453581208}{\million} & \SI{27715835113}{\million} \\
                        & finance & \SI{17525668}{\million}  & \SI{951360680}{\million}   \\
                        & all     & \SI{471106876}{\million} & \SI{28667195793}{\million} \\
        \bottomrule
    \end{tabular}
    \caption{Distribution of the training dataset.}
    \label{tab-data-overview}
\end{table}

We applied temperature sampling (\( t=5 \)) in order to increase the visibility of under represented pairs. Given a collection \( \mathcal{D} \) of datasets, the probability of choosing a sample from a dataset \( D_i \in \mathcal{D} \) after temperature sampling is given by \( P_t(D_i) \) and is calculated from the original dataset statistical distribution \( P(D_i) \).
\begin{align*}
    P(D_i)    & = \frac{N_i}{\sum_{i=0}^{\lvert \mathcal{D} \rvert }{N_i}} \\
    T(D_i, t) & = P(D_i)^{1.0 / t}
\end{align*}
where \( N_i \) is the size of dataset \( D_i \) and \( T(D_i, t) \) is the factor by which the dataset \( D_i \) should be oversampled. The new size \( k_i \) of the oversampled dataset \( D_i \) is given by:
\begin{align*}
    k_i = \left\lfloor \frac{T(D_i, t) \cdot \max_{i=0}^{\lvert \mathcal{D} \rvert}(N_i)}{\max_{i=0}^{\lvert \mathcal{D} \rvert}(T(D_i, t)} \right\rfloor
\end{align*}
Finally, the probability of picking a sample from dataset \( D_i \) after temperature sampling is given by
\begin{equation*}
    P_t(D_i) = \frac{k_i}{\sum_{i=0}^{\lvert D \rvert }{k_i}}
\end{equation*}
Since the balance between general and financial is also extremely skewed, we applied the temperature sampling separately on the general and financial domains.

\subsection{Tokenizer}

As we planned to train a multilingual model, we trained a Byte-Level BPE tokenizer~\citep{wei2021training} from scratch because, according to the authors, it is expected to better share the tokens among the multiple languages, resulting in less rare tokens and, hence, better embeddings. The tokenizer has been trained on the whole, non-oversampled, dataset, and we set the vocabulary size to \num{100000}.

We also reserved a small set of special tokens representing the supported languages and domains. They are inserted inside the input sequence so the model knows this information while generating a translation. For instance, the \textit{English language token} is \texttt{<lang\_en>} and the \textit{general domain token} is \texttt{<dom\_general>}.

\subsection{Data format}

Each sample of the datasets has two categories of features: inputs and outputs.
Input features are data that will be given during inference, and output features are data that should be predicted by the model.
Inputs are:
\begin{itemize}
    \item source sentence;
    \item target language (because the model needs to know the desired target language).
\end{itemize}
Outputs are:
\begin{itemize}
    \item source language;
    \item domain;
    \item translated sentence.
\end{itemize}

Predicting the source language is not required, but we decided to include it to give to the model the ability to automatically detect the source language, as it is a very common and handy feature of most commercial translation tools. One could argue that this should be an input parameter, but we decided that the model should be able to classify by itself the language of the source sentence. Yet, the source language token can still be given as input at inference time to force a particular language. This also apply to the domain token.

Since we plan to train a decoder-only model, training samples have been formatted such that the input tokens are first seen by the model, so the model has access to the whole input when generating the first output token. This is why we chose to encode the sentence pairs in the following format:
\begin{center}
    \texttt{SOURCE </src> <target lang> <source lang> <domain> TARGET <eos>}
\end{center}
where \texttt{</src>} and \texttt{<eos>} are special tokens used to indicate, respectively, the end of the source and target sequences.

This data format gives the possibility to either provide the source language if required, or let the model predict it automatically. For instance, in the real example below, the green part represents the mandatory input (source sentence and target language), the orange part the optional input (source language) and the red part is the output generated by the model.
\begin{center}
    \texttt{%
        \textcolor{ForestGreen}{The buyer pays at an ATM. </src> <lang\_fr>}
        \textcolor{Orange}{<lang\_en>}
        \textcolor{Orange}{<dom\_general>}
        \textcolor{BrickRed}{L'acheteur effectue le paiement sur les bornes automatiques. <eos>}}
\end{center}

\subsubsection{The \texttt{<eos>} token issue}\label{sec-eos-token-issue}

All the models were trained in the same way LLM are trained. Sentence pairs were packed until the training batch was completely filled. These samples were separated by the usual end-of-sentence token \texttt{<eos>}. Ideally, one should also apply proper masking so tokens cannot attend to tokens from past sentence pairs. However, this features is not implemented in flash-attention 2~\citep{dao2022flashattention}, so we trained the models without masks (except the causal mask). We expect the training task to be slightly more complex to solve, as the model now needs to learn to ignore every token before an \texttt{<eos>} token, but we decided that the gain in training speed is worthwhile.

\begin{table}[ht]
    \centering
    \begin{tabular}{lrr}
        \toprule
        Model & without \texttt{<eos>} & with \texttt{<eos>} \\
        \midrule
        70M   & \num{30.80}            & \num{41.11}         \\
        160M  & \num{39.12}            & \num{45.13}         \\
        410M  & \num{40.85}            & \num{46.82}         \\
        \bottomrule
    \end{tabular}
    \caption{BLEU scores of the same models when sources are prefixed with and without the \texttt{<eos>} token.}\label{tab-eos-bleu}
\end{table}

Our initial experiments showed that the quality of translations generated by the models were far below our expectations. We found that the absence of the \texttt{<eos>} token before the source sentence was confusing the model, explaining the drop in translation quality shown in \Cref{tab-eos-bleu}. The \texttt{<eos>} token, which was meant to signal the end of the translation, is actually also interpreted as a \enquote{start of translation} token. Indeed, during training, all sentence pairs (except the first one) are prefixed with the \texttt{<eos>} token. This phenomenon is clear in the example below, in which three sentence pairs are packed in the same training sample.

\begin{center}
    \texttt{%
        \textcolor{JungleGreen}{x\textsubscript{1-1} x\textsubscript{1-2} x\textsubscript{1-3} </src> <tgtlang\textsubscript{1}> <srclang\textsubscript{1}> y\textsubscript{1-1} y\textsubscript{1-2} \textbf{<eos>}}
        \textcolor{Plum}{x\textsubscript{2-1} x\textsubscript{2-2} </src> <tgtlang\textsubscript{2}> <srclang\textsubscript{2}> y\textsubscript{2-1} y\textsubscript{2-2} \textbf{<eos>}}
        \textcolor{Bittersweet}{x\textsubscript{3-1} x\textsubscript{3-2} x\textsubscript{3-3} </src> <tgtlang\textsubscript{3}> <srclang\textsubscript{3}> y\textsubscript{3-1} y\textsubscript{3-2} y\textsubscript{3-3} y\textsubscript{3-4} \textbf{<eos>}}
        \textcolor{Gray}{x\textsubscript{4-1} \ldots}
    }
\end{center}

\begin{figure}[ht]
    \centering
    \includegraphics[width=\linewidth]{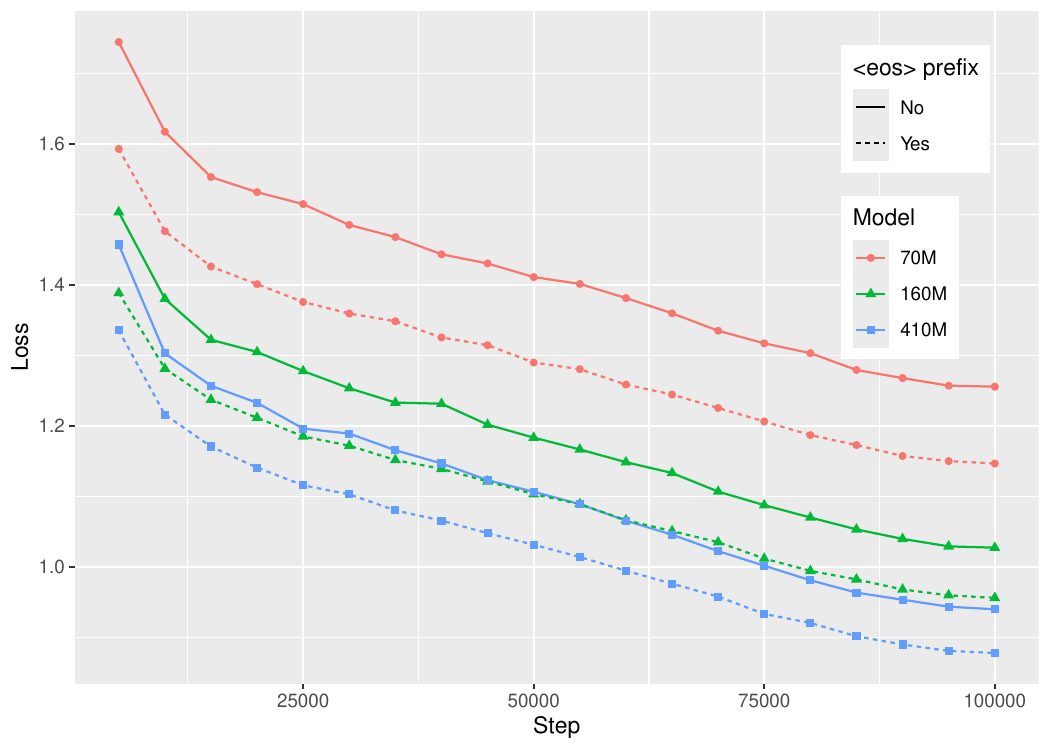}
    \caption{Test loss of our three smallest models (70M, 160M and 410M) with and without the \texttt{<eos>} prefix.}\label{fig-eos-issue}
\end{figure}

The impact of \texttt{<eos>} absence on the test loss can be seen in \Cref{fig-eos-issue}. The model clearly outputs better translation when the source sentence is prefixed with an \texttt{<eos>} token. This is particularly blatant when comparing the 160M and 410M models, respectively with and without the \texttt{<eos>} token prefix. The 410M model, albeit being more than two times bigger than the 160M model, cannot generate better translations without the \texttt{<eos>} prefix.

This problem should be negligible when training LLM, as documents are usually longer than sentence pairs, so \texttt{<eos>} tokens are scarcer. However, its impact will increase as batch size grows, since more sentence pairs can be packed into the same batch. We experimented with a relatively small input length (512 tokens) and the absence of the \texttt{<eos>} token during inference already lead to significant drop in performance. Generally speaking, this issue should not be ignored when more than one sequence are packed in a single training sample. When possible, one should properly mask previous training samples. As it is not possible, currently, to leverage the state-of-the-art self-attention algorithms, we recommend to always prefix all source sentences with the same prefix token(s), both during training and inference. In the remaining of this paper, we will only consider translations generated this way.

\subsection{Training strategy}

As we aim to train models dedicated to the translation task, we computed the loss only on target tokens, so the model learns to generate only text given a source sentence. This is different from pre-trained language models as there is no notion of source and target sentence. The \textit{target-only} strategy has proven to be effective for training text-to-text models~\citep{touvron2023llama}, and is also similar to the way loss of encoder-decoder models is calculated, which are commonly used for machine translation~\citep{costa2022no}. Finally, we packed as many sentence pairs that we could in a single batch, in order to increase the training efficiency.

\subsection{Model architectures}

We used almost the same model architectures used in the Pythia suite~\citep{biderman2023pythia}, the only difference being the number of attention head of the 160M model, as flash-attention expects a multiple of 8. We trained the models using the GPT-NeoX library~\citep{gpt-neox-library}. We made a few changes to the data processing scripts in order to ignore source tokens during the loss computation. An overview of the different models we trained is given in \Cref{tab-model-sizes}.

\begin{table*}[ht]
    \centering
    \begin{tabular}{lrrrrrr}
        \toprule
        Model & Non-embedding    & Embedding       & Layers & Dim  & Heads & Max \( lr \)  \\
        \midrule
        70M   & \num{70295552}   & \num{51380224}  & 6      & 512  & 8     & \( 1e^{-3} \) \\
        160M  & \num{162126336}  & \num{77070336}  & 12     & 768  & 16    & \( 1e^{-3} \) \\
        410M  & \num{405071872}  & \num{102760448} & 24     & 1024 & 16    & \( 1e^{-3} \) \\
        610M  & \num{607448064}  & \num{154140672} & 16     & 1536 & 16    & \( 1e^{-3} \) \\
        1B    & \num{1011257344} & \num{205520896} & 16     & 2048 & 8     & \( 1e^{-4} \) \\
        6.9B  & \num{6855204864} & \num{411041792} & 32     & 4096 & 32    & \( 1e^{-4} \) \\
        \bottomrule
    \end{tabular}
    \caption{Architectures of the trained models. They closely follow the Pythia models but parameters counts do no match because of the bigger vocabulary size, which increases the size of both the embedding and classification layer.}\label{tab-model-sizes}
\end{table*}

All models are trained with a fixed batch size of \num{262144} tokens (512 sequences of length 512 tokens) per GPU, on 8 Nvidia A100 GPUs. The models are trained in bfloat16 precision using the Adam optimizer with weight decay set to \num{0.1}, \num{100} warmup steps and cosine learning rate decay. The maximum learning rate of sub-1B models is set to \num{1e-3}, and \num{1e-4} for larger model because of loss instabilities during the training.

The models are trained for \num{100000} steps, on a total of \num{209715200000} (200B) tokens, although only half of them were actually used to train the model as we do not take into account source tokens when calculating the loss.

\section{Experiments and results}


In this section, we will study the impact of variations in training data size and parameters count on the test loss, for all our models. We will also verify if these changes correlate with their real translation performances using standard metrics such as BLEU and COMET. We finally explore two different model scaling strategies.

\subsection{Applying machine translation scaling law}

\begin{figure}[ht]
    \centering
    \includegraphics[width=\linewidth]{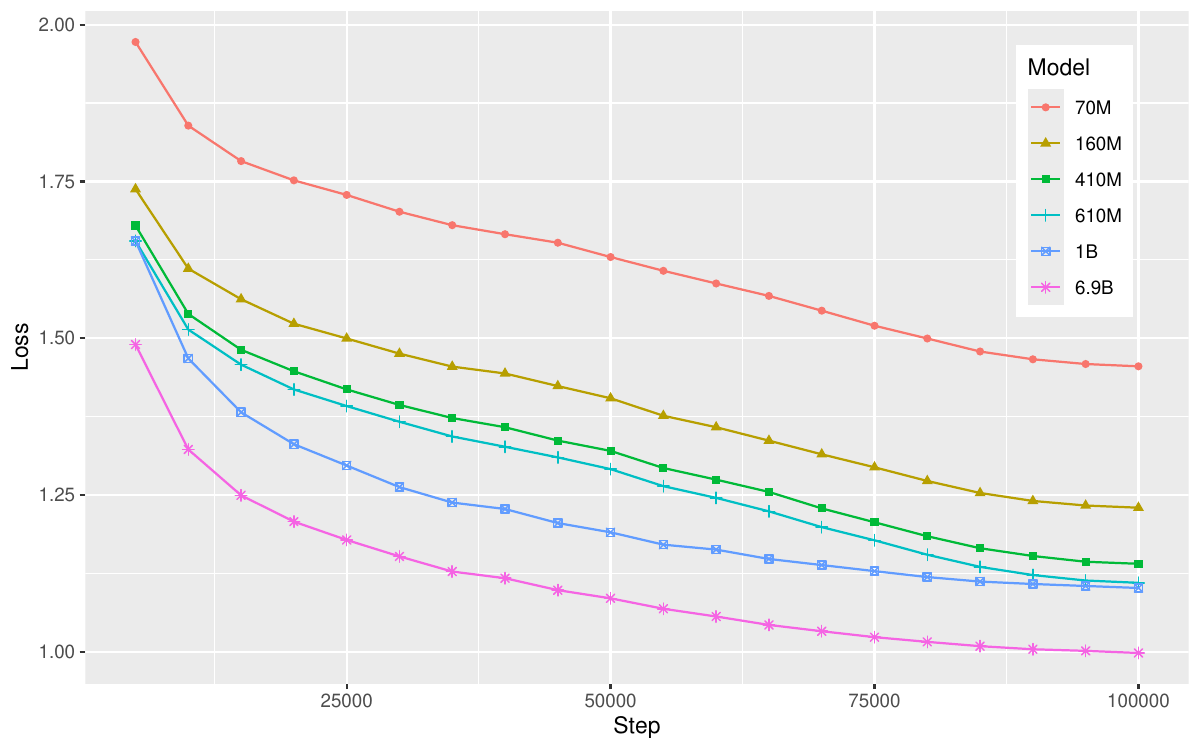}
    \caption{Test loss of all model checkpoints. Each step represents 512 training samples. Larger models always converge faster given the same amount of training data.}\label{fig-loss-per-steps}
\end{figure}

All existing scaling-laws studies show that larger models exhibit better generalization capabilities~\citep{gordon2021data,fernandes2023scaling,ghorbani2021scaling,rae2021scaling,kaplan2020scaling,biderman2023pythia}. This study is no exception, as can be seen in \Cref{fig-loss-per-steps}, larger decoder models always converge faster and require less training data to reach the same loss value.

We first fitted multiple curves following the setting of \citet{ghorbani2021scaling,fernandes2023scaling}, who studied scaling laws for machine translation. The form of the law is given below:
\begin{equation}
    L(N) = \alpha {N^{-p}} + \beta
\end{equation}
where \( N \) is the number of trainable parameters, and the other variables are fitted by minimizing the huber loss (with a delta value of \( 0.01 \)) using the \texttt{BFGS} algorithm from \texttt{SciPy}~\cite{2020SciPy-NMeth}.

\begin{figure}
    \centering
    \includegraphics[width=0.98\linewidth]{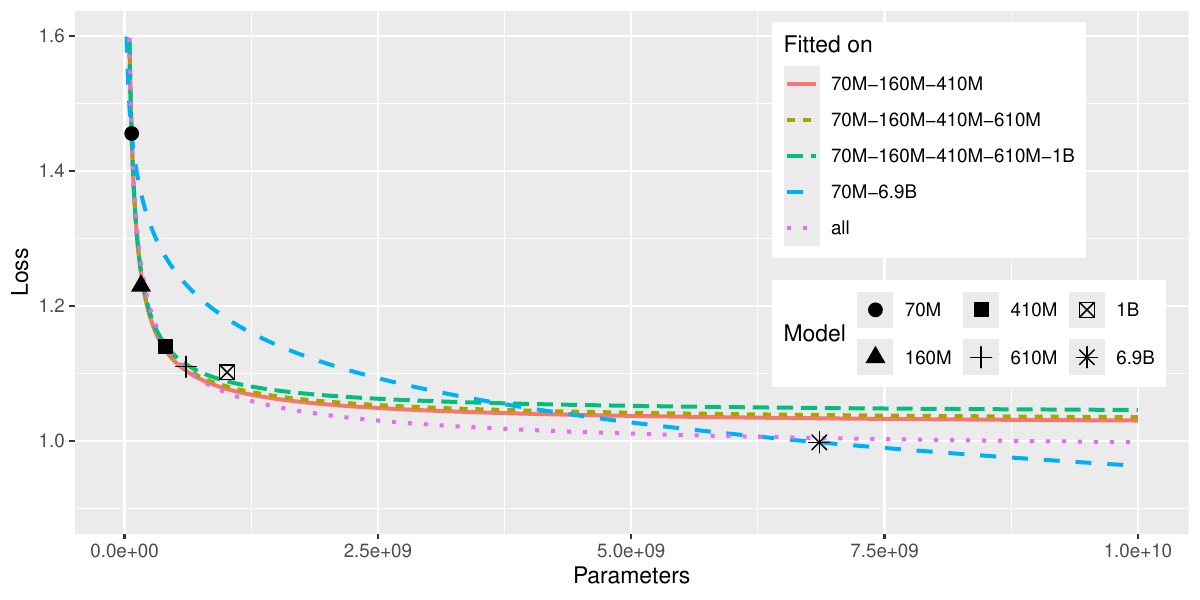}
    \caption{Test losses estimated by power law fitted on different subset of models. Laws fitted on all models and 70M-160M-410M-1B models subset match our observations.}\label{fig-simple-power-law}
\end{figure}

As shown in \Cref{fig-simple-power-law}, the test losses of our translation models can be realistically described by the power law fitted on observations made on all our models (the purple dotted line). This suggests that, indeed, performances of translation models follow a scaling law, that can be expressed by the formula above.
We also fitted curves on less data points in order to verify if we could estimate the loss of the 6.9B model. Unfortunately, the fitted curves become less relevant as soon as we remove the data points from the largest model (the 6.9B model). This is extremely problematic, as the main goal of scaling laws is to estimate the performances of not-yet-trained larger models. Yet, we show that it is difficult to find a good estimation of the 6.9B model's performance without actually training it.
For instance, the law fitted on the observations made on the subset \texttt{70M-160M-410M-610M-1B} (in green) cannot give a good approximation of the \textit{unseen} 6.9B model's performance, and the others are even worse. Therefore, we think one might be particularly cautious when applying such scaling laws to estimate larger models behaviors. Even if our law fitted on all data points seems to be a good estimator of the test loss, we think it will deviate from real observations as the model grows in size.

\begin{figure}
    \centering
    \includegraphics[width=0.98\linewidth]{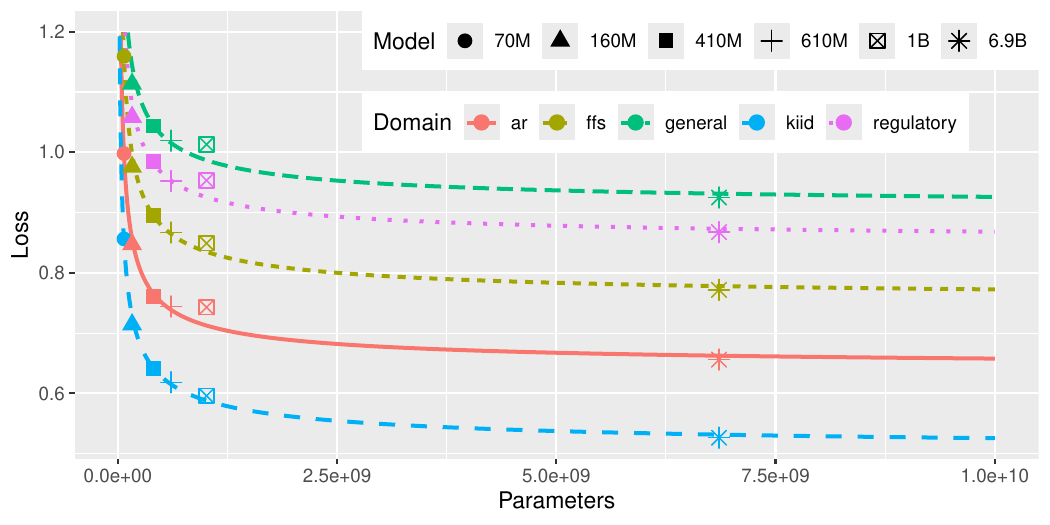}
    \caption{Scaling law fitted on the general domain and some financial subdomains. The law are fitted on the English-French direction only.}\label{fig-simple-power-law-domains}
\end{figure}

We also fitted scaling laws on a per-domain and per-direction basis, on all available data points. This is particularly interesting as it highlights discrepancies between domains and directions. As shown in \Cref{fig-simple-power-law-domains}, it seems to be significantly easier to translate sentences from the \textit{kiid} (Key Investor Information Document) financial domain, but translating general domain sentences is the most difficult, even though the huge majority of our training set is from the general domain. We suspect this curve are, somehow, indicators of the diversity inside each domain. Indeed, \textit{kiid} documents are, by law, all following the same structure and must contain a specific set of information, written in a certain way. On the contrary, \textit{general} domain documents do not follow any rule, making this domain the most heterogeneous one, and thus the most difficult to translate. Other phenomena might explain the differences between these curves. For instance, we also think the presence of many very specific and rare words in the \textit{regulatory} domain explains partly the lower translation quality in this domain.

\begin{figure}
    \centering
    \includegraphics[width=0.98\linewidth]{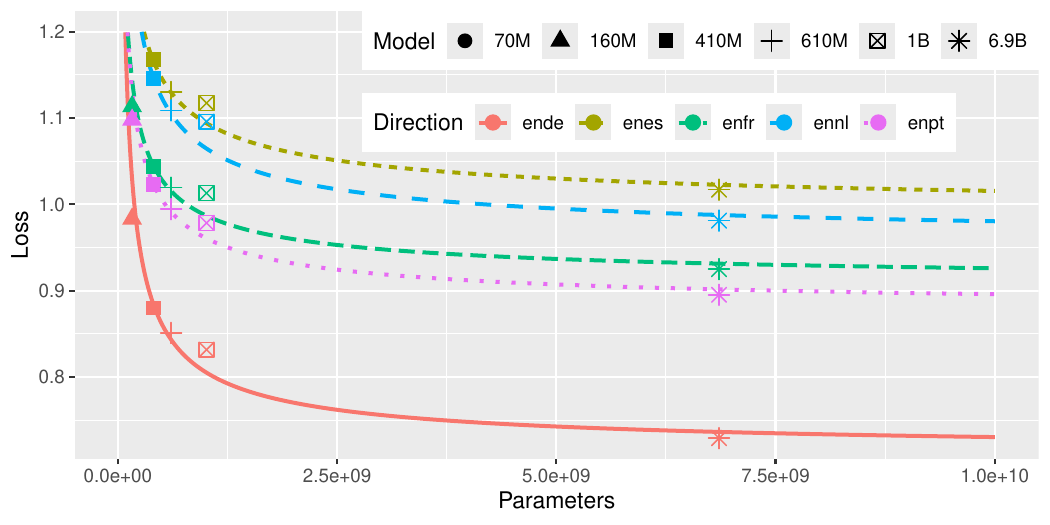}
    \caption{Scaling law fitted on the general domain for all English-X direction.}\label{fig-simple-power-law-en-x}
\end{figure}

We also fitted one curve per direction and observed similar phenomena, as shown in \Cref{fig-simple-power-law-en-x}. For example, our models seem to be better at translating from English to German than from English to French, although our training dataset contains twice as many English-French pairs (before oversampling).

These observations show that the scaling behavior of translation models depends on the training data distribution, and thus scaling laws estimated on a given dataset will not match the real scaling behavior on another one, although they might have the same general shape. For instance, it is not realistic to rely on a scaling law fitted on the EN-FR direction to estimate the performances on the EN-DE direction.

\subsection{Applying language modeling scaling law}

So far, we experimented with a scaling law formula based on the model size only, ignoring the training dataset size. Even if we just showed that lower perplexity/loss can be obtained with fewer data samples (in the case of the en-fr and en-de directions), larger training datasets still tend to increase the overall models' quality. But, it's also a waste of computing resources to train a model on more data than required, this is why modern language modeling scaling formula take into account both the number of trainable parameter and the training dataset size. Hence, we fitted multiple Chinchilla laws following the setting of \citet{hoffmann2022training}, whose form is given below, on various combinations of input data to see if it can be used to reliably predict model performances.
\begin{equation}
    L(N, D) = E + \frac{a}{N^{\alpha}} + \frac{b}{D^{\beta}}
\end{equation}
\( E \), \( a \), \( \alpha \), \( b \) and \( \beta \) are variables fitted by minimizing the huber loss (with a delta value of \( 0.01 \)) using the \texttt{BFGS} algorithm from \texttt{SciPy}~\cite{2020SciPy-NMeth} ; \( N \) and \( D \) are, respectively, the number of non-embedding parameters of the model and the number of training samples. More details are given in the original paper.

\begin{figure}[ht]
    \centering
    \includegraphics[width=\linewidth]{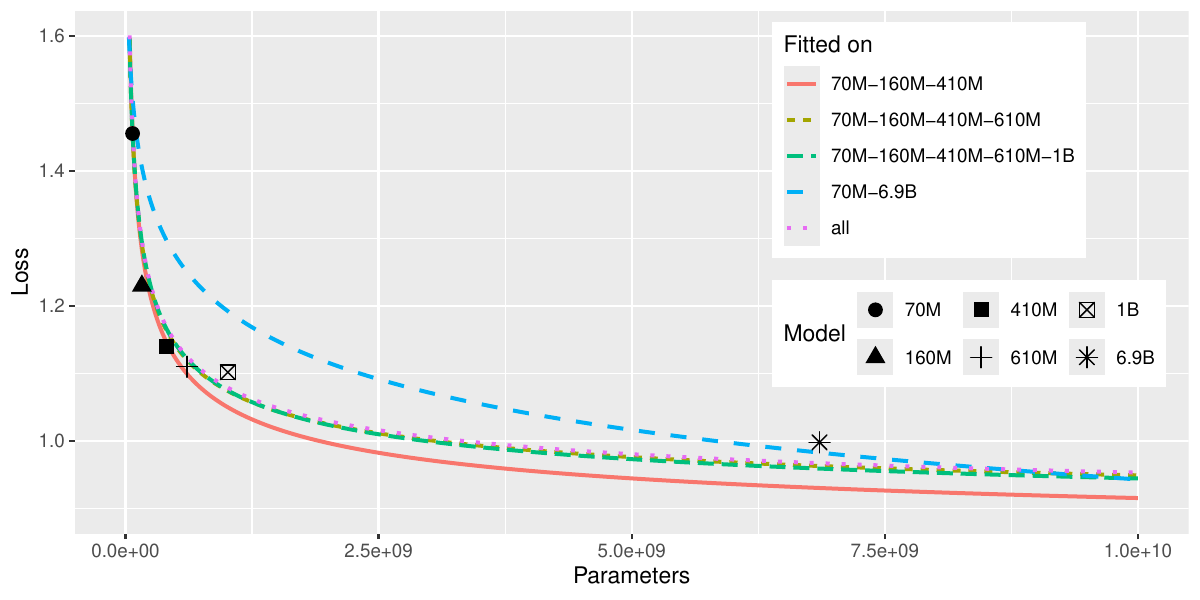}
    \caption{Test losses estimated by the Chinchilla law fitted on different model subsets. Curves deviate from the real observations when we remove too many data points to fit the curve.}\label{fig-power-laws}
\end{figure}

As shown in \Cref{fig-power-laws}, the test loss of our translation models can be realistically described by the power law fitted on observations made on all our models (the purple dotted line). Furthermore, the general shape of the fitted curves is more stable, and thus more trustworthy. Indeed, the curve fitted on all models is very close to the one fitted without the 6.9B model, indicating that behaviors of larger models can be better estimated with this form of scaling law. However, as with the previous scaling law, the curve deviate from real observations when it is fitted on less data points. While it is not a surprising finding, it shows that scaling laws should not be trusted beyond a certain model size. However, we cannot provide a reasonable window in which the estimated loss is realistic.


\begin{figure}[htb]
    \centering
    \includegraphics[width=\linewidth]{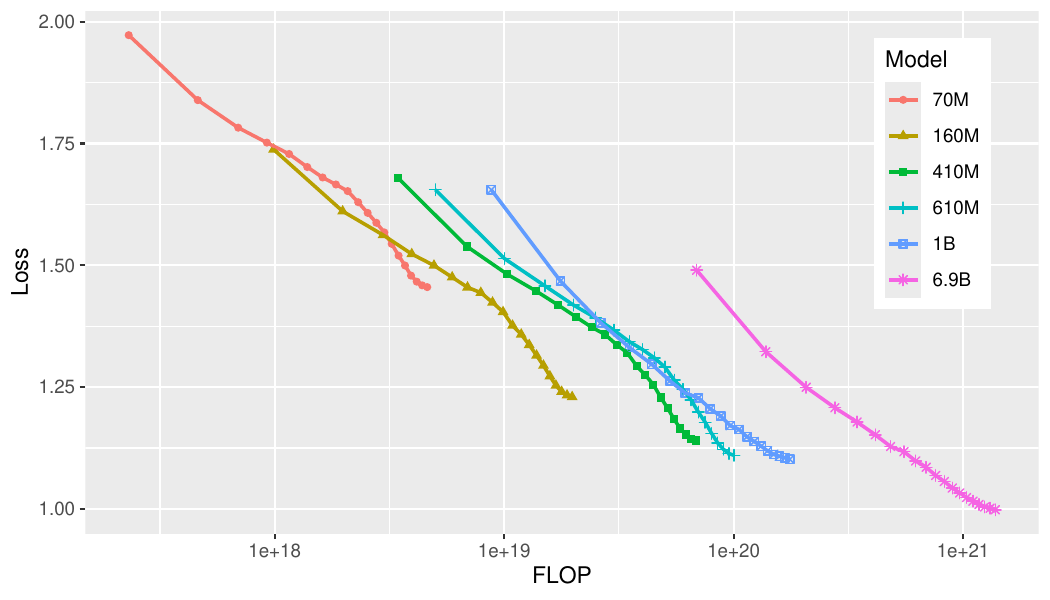}
    \caption{Test loss of all models, each data point represents 5k training steps, or \num{2.5}M samples. Given a fixed FLOP, it's often more beneficial to increase the dataset size when possible.}\label{fig-loss-per-flop}
\end{figure}

\begin{figure}[htb]
    \centering
    \includegraphics[width=\linewidth]{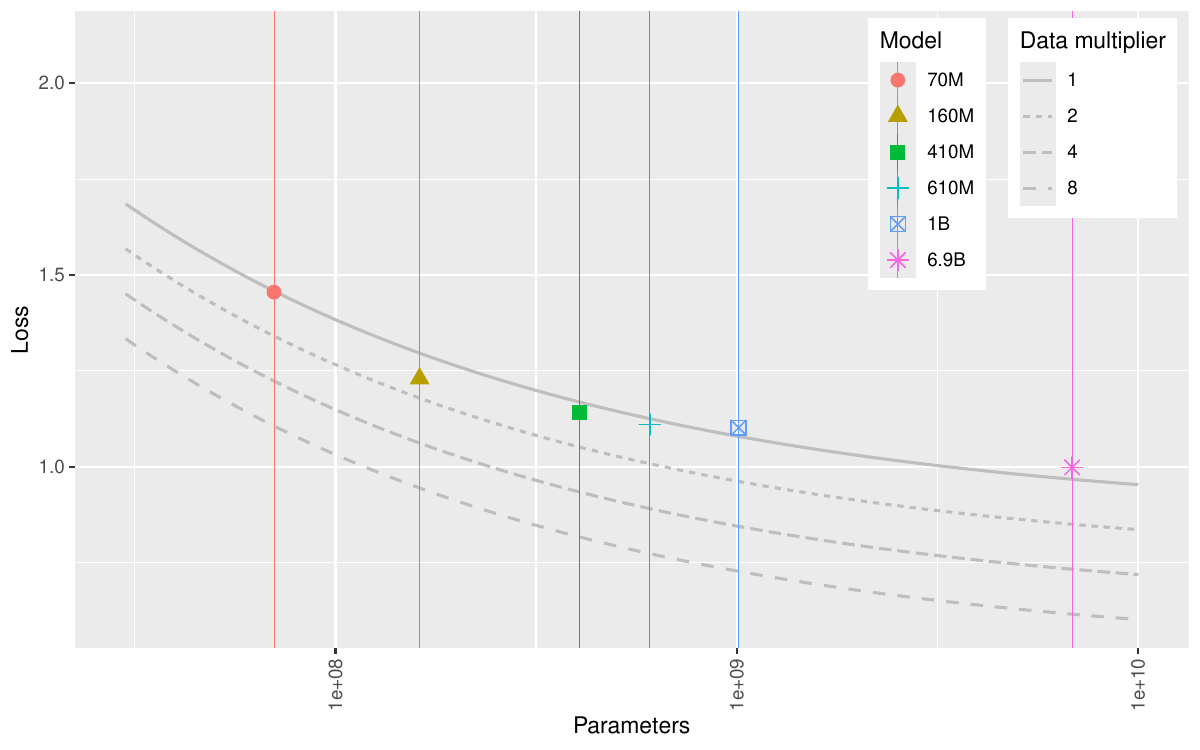}
    \caption{Estimation of models' test losses if they were trained on more data. According to the Chinchilla law fitted on all available observations, the 70M model should be on-par with the 410M performances with four times more data, and the 610M model should match the 6.9B model with only two times more data.}
    \label{fig-more-data}
\end{figure}

These experiments shows two things. First, the test loss of decoder-based translation models follows a scaling law similar to language modeling models, as the curves fitted on all data points matches the real observations. The form of the law (a power law) indicates that larger models will always generalize better, until a certain point where the curve will stay mostly flat. The second thing we show is that finding a good and universal estimation for the model's loss is very difficult, as fitted curves do not generalize well beyond an unknown model size.

\subsection{Correlating scaling law with real translation quality}

\begin{table}[ht]
    \centering
    \small
    \begin{tabular}{llccc}
        \toprule
        Model                       & D & BLEU                                        & COMET                                       & CometKiwi                                   \\
        \midrule
        70M                         & G & \num{29.62}                                 & \num{81.31}                                 & \num{80.72}                                 \\
                                    & F & \num{44.63}                                 & \num{86.95}                                 & \num{80.88}                                 \\
        160M                        & G & \num{32.43}                                 & \num{84.00}                                 & \num{83.45}                                 \\
                                    & F & \num{49.02}                                 & \num{88.27}                                 & \num{81.80}                                 \\
        410M                        & G & \num{33.60}                                 & \num{84.81}                                 & \num{84.14}                                 \\
                                    & F & \num{50.85}                                 & \num{88.64}                                 & \num{81.73}                                 \\
        610M                        & G & \num{34.08}                                 & \num{85.10}                                 & \num{84.35}                                 \\
                                    & F & \num{52.00}                                 & \num{88.85}                                 & \num{81.71}                                 \\
        1B                          & G & \num{34.42}                                 & \num{85.10}                                 & \num{84.33}                                 \\
                                    & F & \num{53.28}                                 & \num{89.98}\textsuperscript{\textdaggerdbl} & \num{81.61}                                 \\
        6.9B                        & G & \num{36.07}\textsuperscript{\textdagger}    & \num{85.88}                                 & \num{84.82}                                 \\
                                    & F & \num{58.34}\textsuperscript{\textdaggerdbl} & \num{89.62}                                 & \num{81.35}                                 \\
        \midrule
        Llama3.1 8B                 & G & \num{30.43}                                 & \num{84.82}                                 & \num{84.47}                                 \\
                                    & F & \num{34.99}                                 & \num{84.42}                                 & \num{81.75}                                 \\
        Tower 7B                    & G & \num{33.50}                                 & \num{85.91}\textsuperscript{\textdagger}    & \num{85.02}\textsuperscript{\textdagger}    \\
                                    & F & \num{38.93}                                 & \num{86.49}                                 & \num{82.66}\textsuperscript{\textdaggerdbl} \\
        Mistral 7B                  & G & \num{23.26}                                 & \num{80.08}                                 & \num{82.29}                                 \\
                                    & F & \num{38.93}                                 & \num{76.52}                                 & \num{76.17}                                 \\
        \midrule
        Tower 7B\textsuperscript{*} & G & \num{34.38}                                 & \num{86.22}                                 & \num{85.23}                                 \\
                                    & F & \num{39.08}                                 & \num{86.52}                                 & \num{82.74}                                 \\
        \bottomrule
    \end{tabular}
    \caption{Evaluation of the six models trained during this study. We reports both the scores on the general (G) domain and average over all financial (F) subdomains. We also include best performing LLM. As Tower has not been trained on Swedish, we also evaluate it after removing directions including Swedish (the Tower 7B\textsuperscript{*} rows). Best scores on the general and financial domains are indicated by \textsuperscript{\textdagger} and \textsuperscript{\textdaggerdbl} respectively.}\label{tab-scores-bleu-comet}
\end{table}

Let us suppose we know the function modeling the real loss given a model size and an amount of training data. We still do not know if targeting lower loss values will actually improve the quality of the translations generated by the model. We provide in the following an empirical study showing the correlation between the model's loss and its translation performance. We computed BLEU~\citep{papineni2002bleu}, COMET~\citep{rei2022comet} and CometKiwi~\citep{rei2022cometkiwi} scores for all six models, and we observed that, indeed, a lower loss does correlate with a performance increase, as shown in \Cref{tab-scores-bleu-comet}. This trend can be observed on the general domain for all directions, as shown in \Cref{sec-model-perf-per-dir}. However, on the financial domain, CometKiwi does not always increase, it reaches a peak on the 610M model, then decreases. We conjecture that CometKiwi cannot correctly evaluate domain specific translations, as it is a reference-free model trained mainly on generalist sentences. We show in \Cref{sec-model-perf-per-dir} that BLEU and COMET always increase with models' size, while CometKiwi often decreases at some point.

We also compare our models to well established LLM, and we show that smaller but specialized models clearly outperforms large and generalist LLM, as shown by our 410M model performing on par with \textbf{Llama 8B}. Our largest models are also real competitors to \texttt{Tower 7B}, even though it has been trained on much more data and specialized for machine translation. \texttt{Tower 7B} has the highest CometKiwi score, we just show, it might not be reliable on specialized domains. Our models are obviously performing better on the financial domain, because only our models were finetuned on financial data. We also remark that Mistral's scores are quite low on the general domain, a quick manual inspection revealed that the model often give details and explanations about the produced translation, even when asked not to. As a consequence, we think that Mistral lower score is mostly caused by the model not following rigorously the instructions.

So, while it certainly boost performances, increasing the model size is often not the optimal solution to improve the model's performance. The training dataset is also extremely important. Indeed, as can be observed in \Cref{fig-loss-per-flop}, given a fixed FLOP budget, it is often preferable to increase the number of training samples. For instance, the 160M model appears to always be better than the 410M, 610M and 1B models given the same FLOP budget, as indicated by the 160M's curve being below other models' curves.
This observation is also validated by the fitted law, as indicated in \Cref{fig-more-data}. Most of the time, and according to the fitted Chinchilla law, it would have been better to just train our models on more data, instead of training larger models. For instance, we estimate that the 160M model would be on-par with the 410M model if trained on approximately twice as many data, which would not exceed the total number of FLOP of our current 410M model.

\begin{table*}[tb]
    \centering
    \begin{tabular}{lrrrrrr}
        \toprule
        Model     & Layers & Dim  & Non-embedding   & Embedding       & FLOP per s.   & Samples per s. \\
        \midrule
        70M       & 6      & 512  & \num{70295552}  & \num{51380224}  & \num{1.06e14} & 1170           \\
        \addlinespace
        70M+d768  & 6      & 768  & \num{119599104} & \num{77070336}  & \num{1.74e14} & 900            \\
        70M+12l   & 12     & 512  & \num{178339840} & \num{51380224}  & \num{1.37e14} & 760            \\
        \addlinespace
        70M+d1024 & 6      & 1024 & \num{178339840} & \num{102760448} & \num{2.43e14} & 725            \\
        70M+24l   & 24     & 512  & \num{127038464} & \num{51380224}  & \num{1.6 e14} & 445            \\
        \bottomrule
    \end{tabular}
    \caption{Sizes and architecture of models scaled in depth (70M+12l and 70M+24l) and models scaled in width (70M+d768 and 70M+d1024) compared to the base 70M model. Increasing the depth of the model has limited impact on the total parameters count, but decreases significantly the efficiency (higher FLOP per second but less samples per second). Scaling the width of the model takes advantage of modern GPU architectures, but adds many trainable parameters.}\label{tab-scaling-efficiency}
\end{table*}

To conclude with, we find that scaling laws are a powerful tool to have a glimpse of what we can expect from a \textit{relatively larger} model trained on the same dataset, but it will probably fail to predict the performances of \textit{much larger} models, even if trained on a similar data distribution. 
It has to be kept in mind when using such scaling laws to plan a training budget: \textbf{at some point, the fitted law will fail}. Planning a training budget based on observations made on a 10B model might be fine to train a 70B model, but completely wrong for a 500B one. Furthermore, a given scaling law can only estimate the end performances of a model trained on the same data distribution used to fit the scaling law. For instance, we show in \Cref{fig-simple-power-law-en-x,fig-simple-power-law-domains} that laws fitted on different language directions or domains are very different, and thus should not be applied to estimate the performances of the model on another direction.

\subsection{Scaling strategies}

\begin{figure}[htb]
    \centering
    \begin{subfigure}{.5\linewidth}
        \centering
        \includegraphics[width=\linewidth]{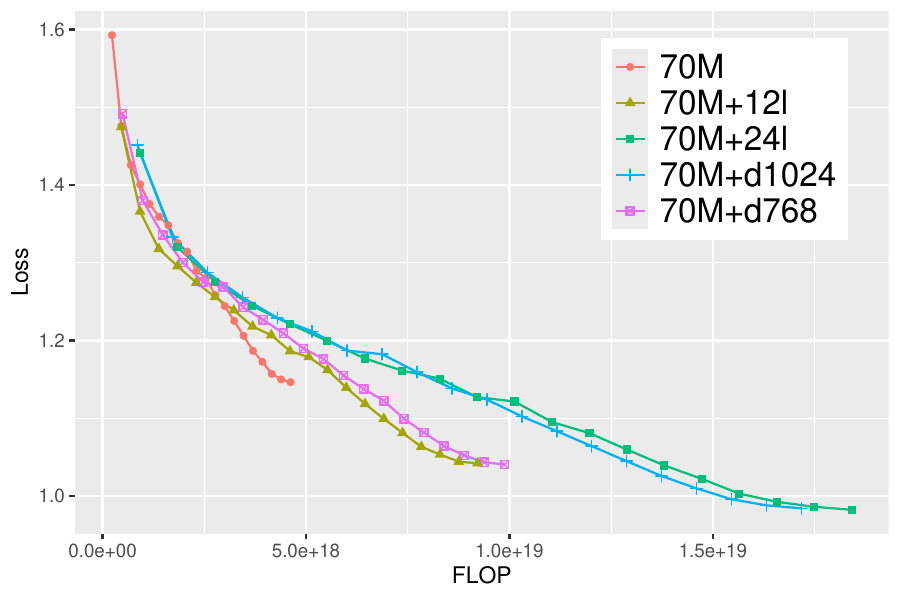}
    \end{subfigure}%
    \begin{subfigure}{.5\linewidth}
        \centering
        \includegraphics[width=\linewidth]{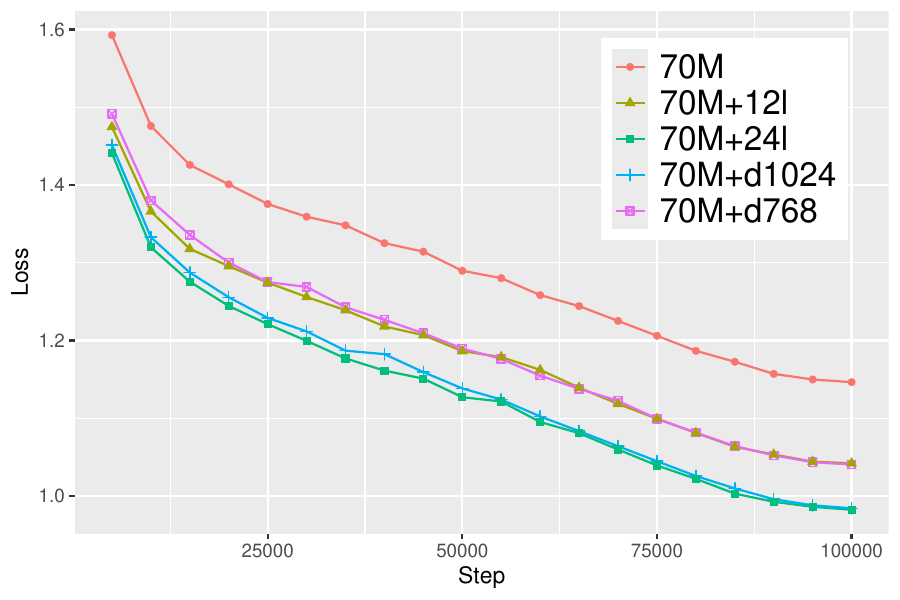}
    \end{subfigure}
    \caption{In our experiments, we increased the width and the depth of the 70M model so the additional cost in terms of FLOP is similar (left). Scaling the depth or the width can lead to similar performance gains (right). The two figures are similar, except that the loss decrease can be observed either through the FLOP budget prism (left) or throughout training time / size of dataset (right).}\label{fig-scaling-deep-vs-wide}
\end{figure}


We also studied whether one should favor scaling the depth (increasing the number of layers) or the width (increasing the hidden size) of a decoder model. We took the smallest model as a baseline and scaled it depth-wise and width-wise so that the increase in parameters increased the total training FLOP by a similar amount, as illustrated in \Cref{fig-scaling-deep-vs-wide}. An overview of the scaled model architectures can be seen in \Cref{tab-scaling-efficiency}. Interestingly, we observed that both scaling methods yield the same performance improvement. As shown in \Cref{fig-scaling-deep-vs-wide}, given a similar FLOP cost, scaling the depth or the width seems to have the very same impact on the test loss.

Generally speaking, scaling depth-wise lead to smaller, but less efficient models. Indeed, modern hardware architecture can handle more efficiently large matrix products than many smaller matrix products. As shown in \Cref{tab-scaling-efficiency}, width-scaled models can generate much more samples than depth-scaled models because the GPU can do more FLOP per second.



\section{Conclusion}

This work describes the behavior of decoder models on the multilingual multidomain machine translation task. We trained models whose number of parameters range from 70M to 6.9B on sentence pairs in eight European languages. We show that decoder-only models for translation tend to scale similarly as language models, as the Chinchilla law can also be applied to our models. As such, we recommend to train machine translation models using the same training recipes as large language models. While we think it is true for most, if not all, NLP tasks, more work need to be carried out to validate this hypothesis. We also highlight a critical limitation of scaling laws: they cannot generalize well beyond an unknown model and/or training dataset size. As models tend to be larger through time, it will be extremely important to find ways to detect early unreasonable deviations of the \enquote{reference} scaling laws on which larger models are build.

We also show that models scaled width-wise appear to be more FLOP efficient than models scaled depth-wise, while reaching almost the same loss. Our experiments need to be continued in order to see when increasing the depth of the model starts to be more valuable than increasing its width. But, generally speaking, increasing the linearly both the depth and the width seems to be a good trade-off between efficiency and parameters count.

Efficient training requires packing as many sentence pairs as possible in a training batch. We discovered that unexpected biases can be introduced if proper masking is not applied, that is to say, if sequences can attend to previous ones. Since it is not possible with current state-of-the-art optimization methods, one must carefully format the training input data. We suggest dropping the \textit{end-of-sentence} token, commonly used to signal the end of text generation, in favor of a \textit{start-of-translation} token signaling the start of a new source sentence and, therefore, the end of the generated target sentence.

This study has been conducted on sentence-level pairs only. While this setup is a bit outdated, it is still the first time a comprehensive study has been made on multilingual machine translation using decoder-only architectures. Nevertheless, we expect decoder models to be easy to adapt to the document-level translation task, as one can simply finetune a sentence-level decoder with non-shuffled sentence pairs from a corpus of parallel documents.

\section{Acknowledgment}

This project was provided with computer and storage resources by GENCI at IDRIS thanks to the grant 2023-AD011014445 on the supercomputer Jean Zay's V100 and A100 partitions.


\bibliography{references}
\bibliographystyle{acl_natbib}

\appendix

\section{Full data distribution}\label{sec-full-data-distribution}

Our models were trained on 11 language directions and 9 domains (8 are financial subdomains + general domain). The list 8 financial subdomains are given below:
\begin{description}
    \item[am] Asset Management
    \item[ar] Annual Report
    \item[corporateAction] Corporate Action Document
    \item[equi] Equity Research
    \item[ffs] Fund Fact Sheet
    \item[kiid] Key Investor Information Document
    \item[lifeInsurance] Life Insurance Document
    \item[regulatory] Regulatory Document
\end{description}

\section{Models' performances per direction}\label{sec-model-perf-per-dir}

\begin{figure}[htb]
    \setlength{\fboxsep}{0pt}
    \setlength{\fboxrule}{0.1pt}

    \centering
    \begin{subfigure}{\linewidth}
        \fbox{\includegraphics[width=0.96\linewidth]{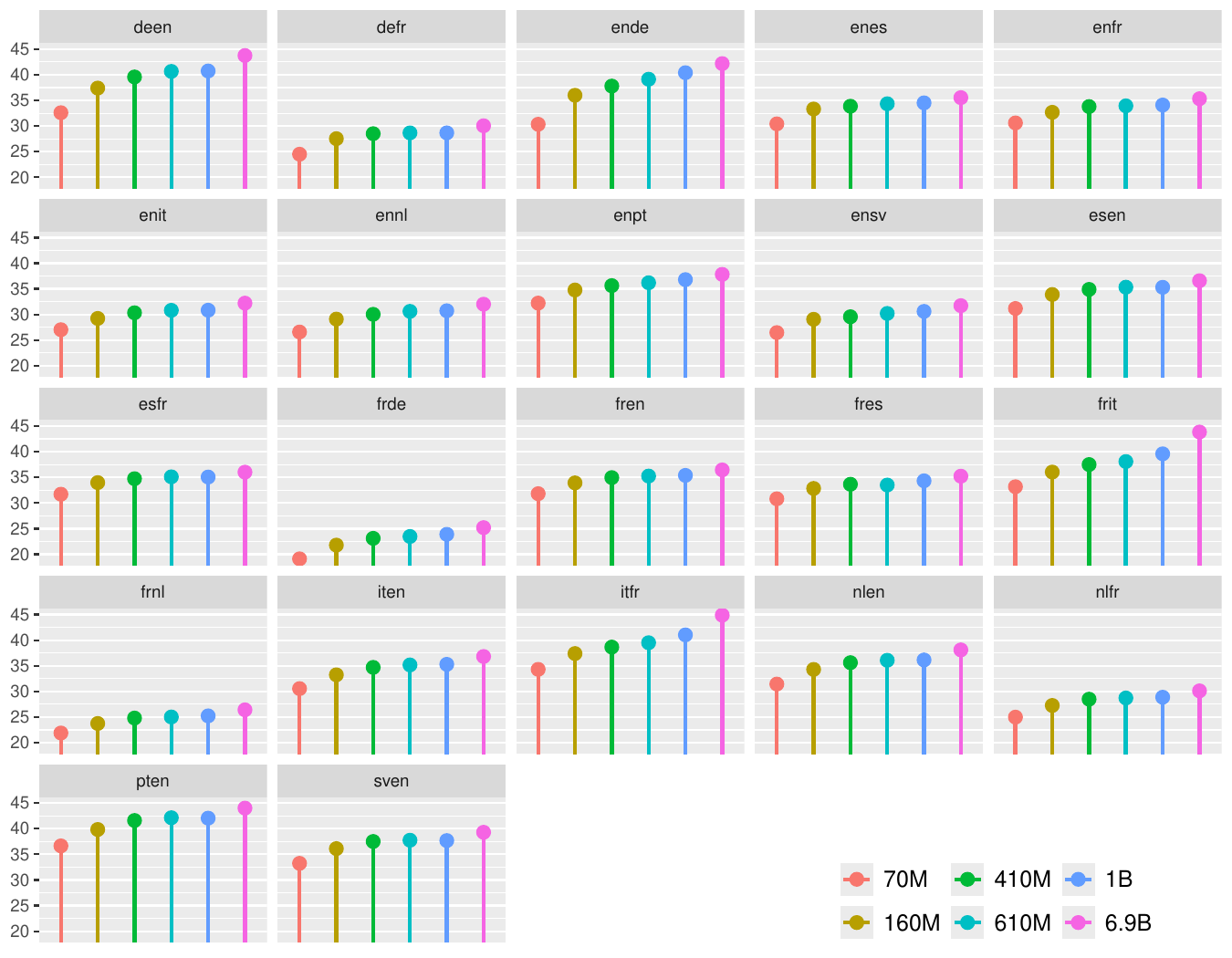}}
    \end{subfigure}

    \smallskip{}

    \begin{subfigure}{\linewidth}
        \fbox{\includegraphics[width=0.96\linewidth]{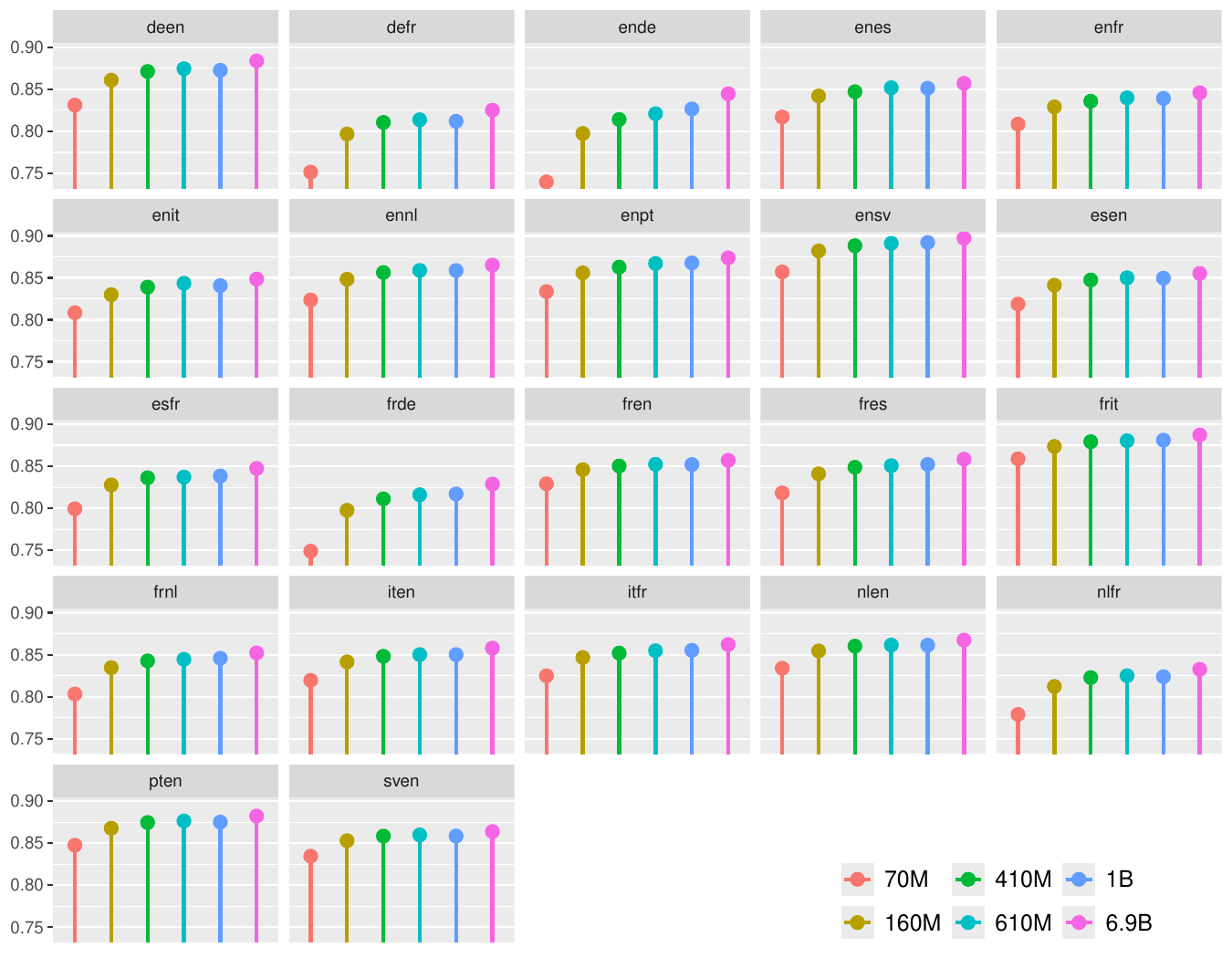}}
    \end{subfigure}

    \smallskip{}

    \begin{subfigure}{\linewidth}
        \fbox{\includegraphics[width=0.96\linewidth]{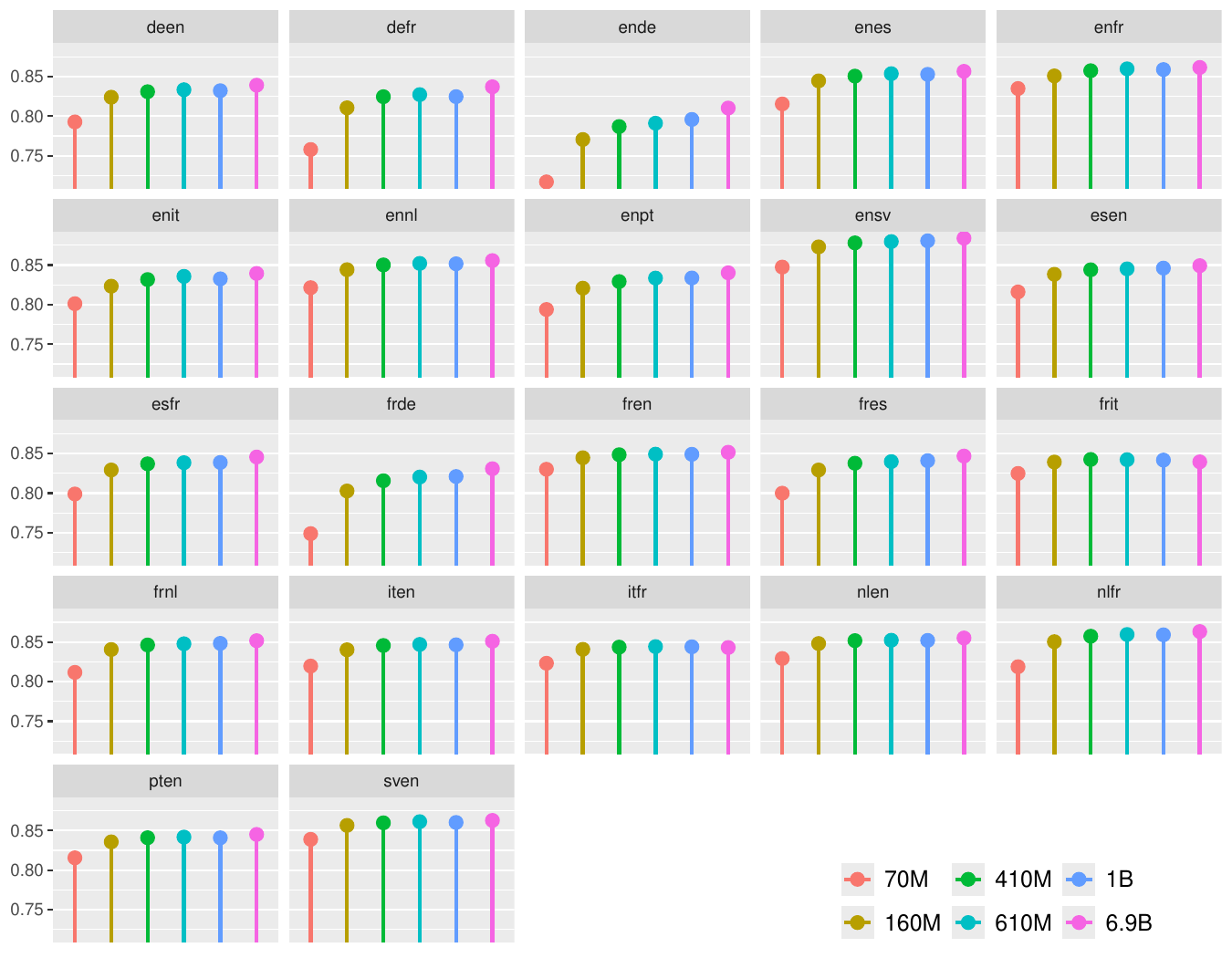}}
    \end{subfigure}

    \caption{From top to bottom, BLEU, COMET and CometKiwi scores computed on the test dataset for all models and directions, on the general domain.}\label{fig-scores-general}
\end{figure}

\begin{figure}[htb]
    \setlength{\fboxsep}{0pt}
    \setlength{\fboxrule}{0.1pt}

    \centering
    \begin{subfigure}{\linewidth}
        \fbox{\includegraphics[width=0.96\linewidth]{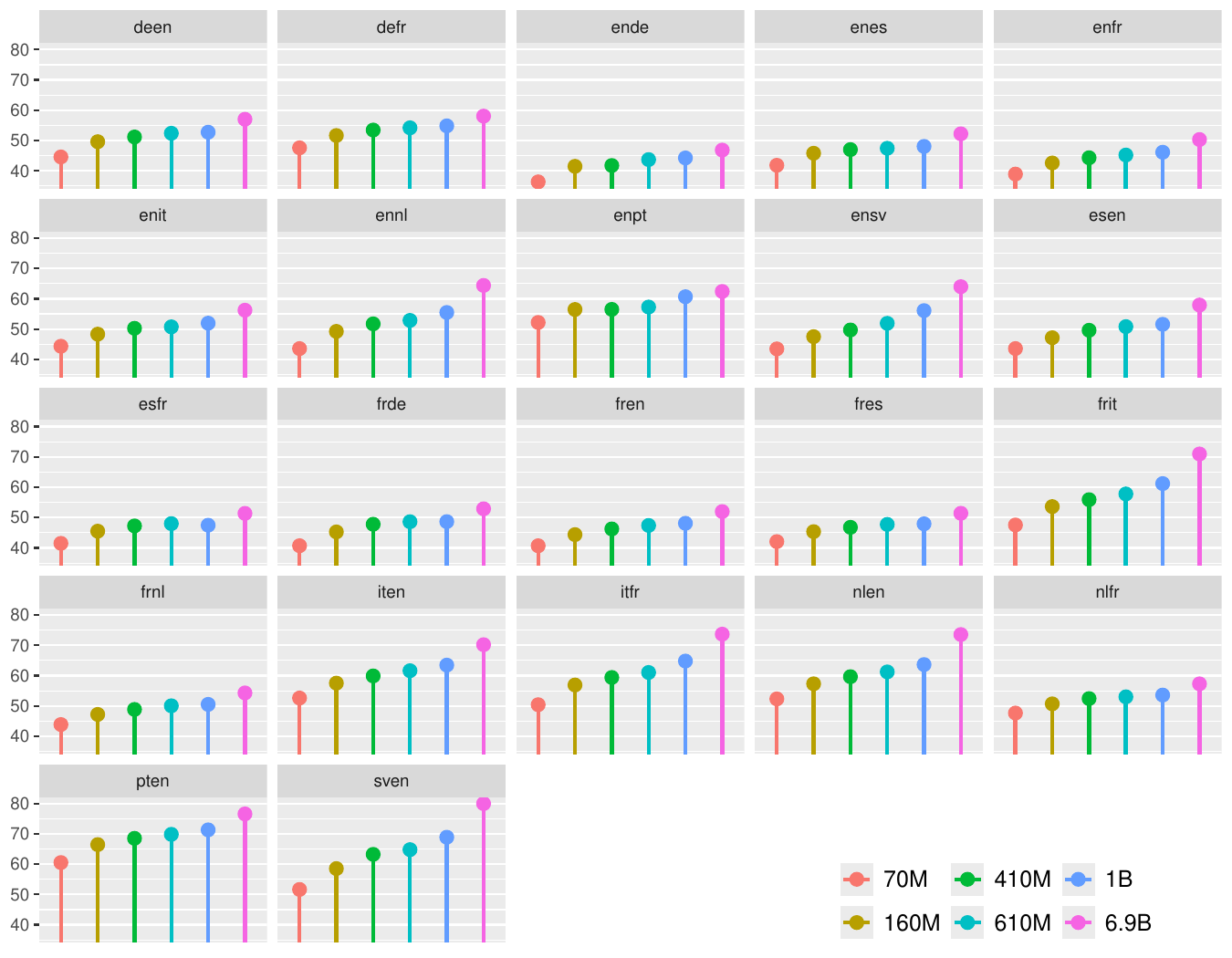}}
    \end{subfigure}

    \smallskip{}

    \begin{subfigure}{\linewidth}
        \fbox{\includegraphics[width=0.96\linewidth]{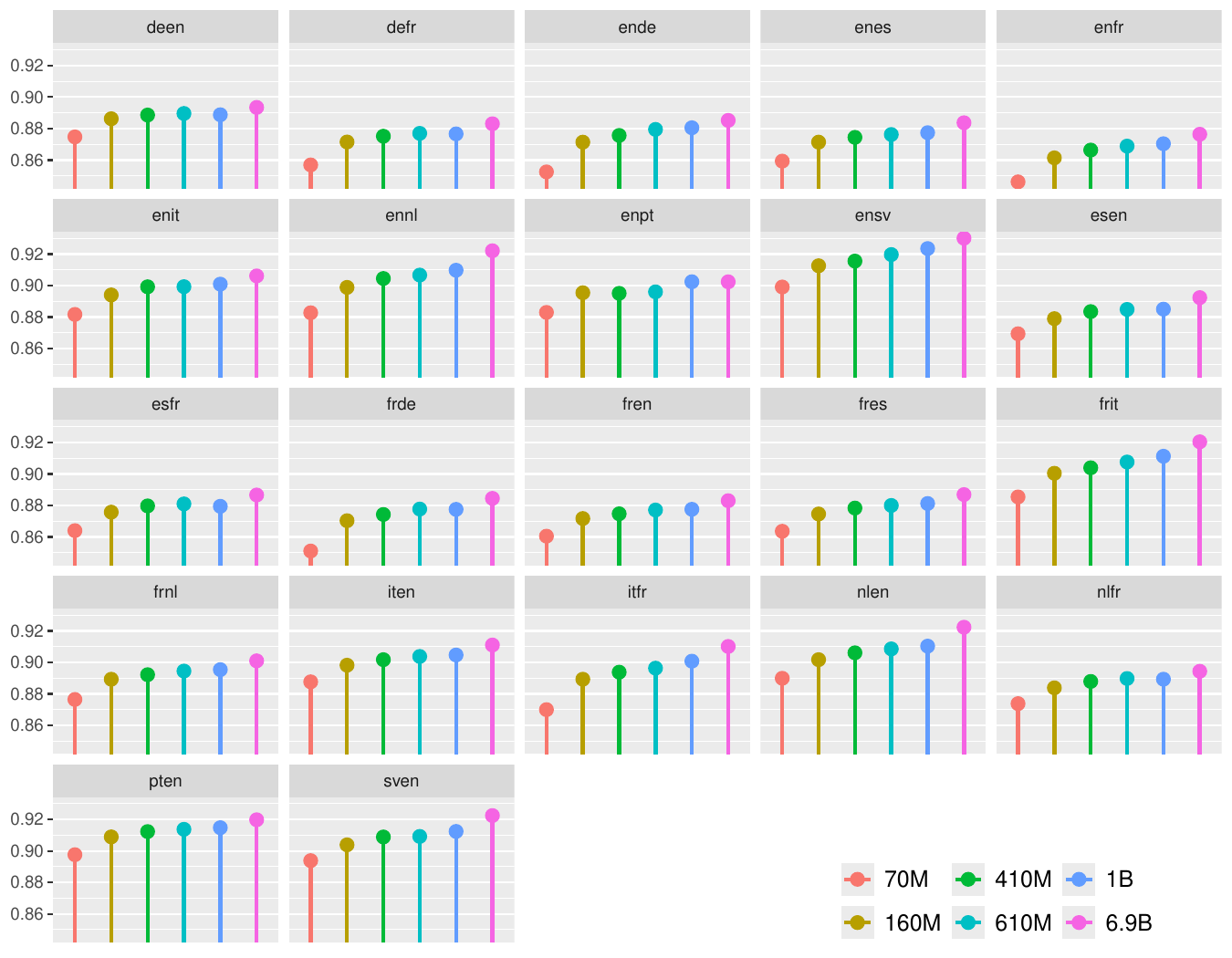}}
    \end{subfigure}

    \smallskip{}

    \begin{subfigure}{\linewidth}
        \fbox{\includegraphics[width=0.96\linewidth]{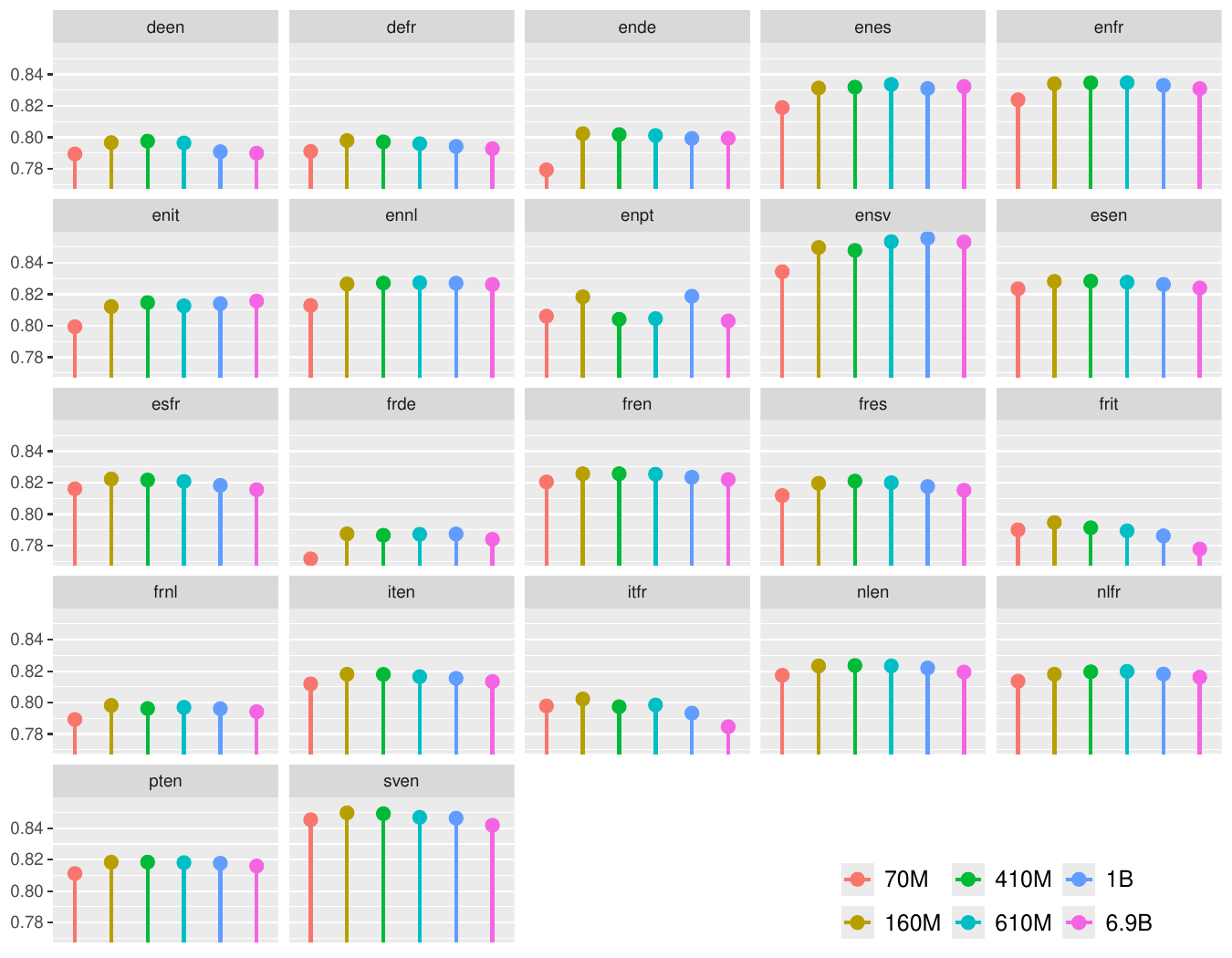}}
    \end{subfigure}

    \caption{From top to bottom, BLEU, COMET and CometKiwi scores computed on the test dataset for all models and directions, averaged over all financial sub-domains.}\label{fig-scores-finance}
\end{figure}

Performances of all models increase as parameters counts increase, regardless of the scoring method, as shown in \Cref{fig-scores-general,fig-scores-finance}. 

\end{document}